\documentclass[runningheads]{llncs}
\usepackage[T1]{fontenc}
\usepackage{graphicx}
\usepackage{bm}
\usepackage{amsmath}
\usepackage{amsfonts}
\usepackage{hyperref}
\usepackage{graphics}
\usepackage{tabularx}
\usepackage{booktabs}
\usepackage{color}
\usepackage{multirow}
\usepackage{siunitx}
\usepackage{appendix}

\begin{document}
\title{Improving Instance Optimization in Deformable Image Registration with Gradient Projection}
\titlerunning{Improving IO in DIR with Gradient Projection}
%
\author{Yi Zhang\inst{1}
\and  Yidong Zhao \inst{1}
\and Qian Tao\inst{1}
}
%
\authorrunning{Y. Zhang et al.}
\institute{Department of Imaging Physics, Delft University of Technology, The Netherlands
\email{\{y.zhang-43, y.zhao-8, q.tao\}@tudelft.nl}}
\maketitle              
\begin{abstract}
Deformable image registration is inherently a multi-objective optimization (MOO) problem, requiring a delicate balance between image similarity and deformation regularity. These conflicting objectives often lead to poor optimization outcomes, such as being trapped in unsatisfactory local minima or experiencing slow convergence. Deep learning methods have recently gained popularity in this domain due to their efficiency in processing large datasets and achieving high accuracy. However, they often underperform during test time compared to traditional optimization techniques, which further explore iterative, instance-specific gradient-based optimization. This performance gap is more pronounced when a distribution shift between training and test data exists. To address this issue, we focus on the instance optimization (IO) paradigm, which involves additional optimization for test-time instances based on a pre-trained model. IO effectively combines the generalization capabilities of deep learning with the fine-tuning advantages of instance-specific optimization. Within this framework, we emphasize the use of gradient projection to mitigate conflicting updates in MOO. This technique projects conflicting gradients into a common space, better aligning the dual objectives and enhancing optimization stability. We validate our method using a state-of-the-art foundation model on the 3D Brain inter-subject registration task (LUMIR) from the Learn2Reg 2024 Challenge. Our results show significant improvements over standard gradient descent, leading to more accurate and reliable registration results. \begingroup
\renewcommand\thefootnote{}
\footnotetext{This paper was prepared as part of the Learn2Reg Challenge of MICCAI 2024.}

\addtocounter{footnote}{-1}
\endgroup
\end{abstract}

\section{Introduction}
Medical image registration establishes anatomical correspondences between two or more medical images. It has become a prerequisite step for many applications, such as treatment planning \cite{staring2009registration,king2010registering}, and longitudinal patient studies \cite{sotiras2013deformable,jin2021predicting}. Traditionally, registration is solved by iteratively solving an optimization problem iteratively \textit{w.r.t.} a parameterized transformation \cite{klein2007evaluation} and a gradient-based optimization algorithm. The optimization objective generally comprises a similarity term to align the images and a regularization term to maintain smoothness. However, the optimization is highly complicated due to the nature that the similarity and regularization can be conflicting, making image registration a multi-objective optimization (MOO) problem \cite{alderliesten2015getting,andreadis2023morea,grewal2024multi}.

Recent advances in machine learning have popularized the data-driven deep-learning paradigm in medical image registration \cite{rueckert2019model}. Traditional optimization methods iteratively update the transformation parameters. In contrast, deep learning-based approaches provide rapid image-to-transformation predictions during inference. The works in the unsupervised learning paradigm employ loss functions similar to those in conventional techniques but optimize them through amortized neural networks \cite{balakrishnan2019voxelmorph,de2019deep}. The potential of deep learning models for medical image registration is evident. Since the emergence of U-Net~\cite{ronneberger2015unet}, a rich line of unsupervised registration networks have been proposed~\cite{balakrishnan2019voxelmorph,dalca2019unsupervised}. Recent methods explored other backbones including transformers \cite{zhang2021learning} or implicit neural representations \cite{wolterink2022implicit,van2023deformable}. 

The learned deep learning-based registration models can be further improved using instance optimization (IO) by fine-tuning the model parameters given a specific test image pair. Due to its performance gain and the affordable extra computational effort, IO has been gaining attention in the community ~\cite{balakrishnan2019voxelmorph,mok2023deformable,zhang2024pca,tian2024unigradicon}. However, the optimal choice of balancing similarity and regularization may differ for each pair in the test time \cite{andreadis2023morea,grewal2024multi}, making the actual gain from IO marginal. With the emergence of large foundation models of image registration \cite{tian2024unigradicon}, the necessity of study an efficient and robust IO method is needed. 

In this work, we introduce a novel instance optimization (IO) algorithm that addresses the multi-objective challenges in image registration. We apply gradient projection techniques, as inspired by advances in MOO \cite{yu2020gradient,dou2023gsmorph}, to enhance the effectiveness of IO. Our method is evaluated using a state-of-the-art foundation model and compared against direct inference and naive IO approaches, specifically for the LUMIR task in the Learn2Reg 2024 Challenge. The experimental results show that our approach significantly outperforms the standard IO method, leading to more accurate and reliable registration results. These findings suggest that gradient projection can be an effective strategy for improving IO in the context of medical image registration.

\section{Methods and Materials}
\subsection{Deformable Image Registration}
In this paper, we consider the deformable image registration given a pair of 3D images $I_{\text{A}} \in \mathbb{R}^{D\times H\times W}$ and $I_{\text{B}} \in \mathbb{R}^{D\times H\times W}$. Our aim is to find a dense transformation $\phi \in \mathbb{R}^{3 \times D\times H\times W} $, such that the warped moving image $I_{\text{A}} \circ \phi$ is anatomically similar to $I_{\text{B}}$. Since in deformable image registration, the magnitude of the transformation is often relatively small, compared to the original image grid $x$, thus denoted by $\phi = x + u(x)$. In general, finding such a deformation field $\hat{\phi}$ can be regarded as a multi-objective optimization (MOO) problem given by \begin{equation}
\hat{{\phi}}=\underset{{\phi}}{\operatorname{argmin}} \ \mathcal{L}_{\text {sim }}(I_{\text{A}} \circ {\phi}, I_{\text{B}})+\lambda \mathcal{L}_{\text {reg }}({\phi}),
\label{eq: phi optimization}
\end{equation}
where $\mathcal{L}_\text{sim}$ is a similarity term between $I_{\text{A}} \circ {\phi}$ and $I_{\text{B}}$, $\mathcal{L}_\text{reg}$ is a regularization on ${\phi}$, and $\lambda$ is a trade-off weight term. In the MOO paradigm, the use of $\lambda$ as a weight for balancing $\mathcal{L}_{\text{sim}}$ and $\mathcal{L}_{\text{reg}}$ is called linear scalarization.

\subsection{Model Architecture}
We use the publicly-available foundation model for pairwise image registration, uniGradICON~\cite{tian2024unigradicon}, which is based on the GradICON registration network~\cite{tian2023gradicon}. The images were passed to a cascaded composition of four U-Nets~\cite{ronneberger2015unet} on three resolution levels with the full resolution repeated twice, resulting in a resolution of $[1/4,1/2,1,1]$. A more detailed architecture can be found in the original paper of GradICON.

\subsection{Loss Functions}
In this work, we use local normalized cross-correlation (LNCC) with a Gaussian kernel of 5 voxels~\cite{vishnevskiy2016isotropic} as $\mathcal{L}_{\text{sim}}$, but symmetrically, \textit{i.e.}, the roles of images in a pair are swapped for inverse registration prediction. We follow the formulation of the gradient inverse consistency regularizer proposed in GradICON. The overall loss is defined by:
\begin{equation}
    \mathcal{L} = \mathcal{L}_{\text{sim}}(I_{\text{A}} \circ {\phi}_{\text{mov}}, I_{\text{B}})+\mathcal{L}_{\text{sim}}(I_{\text{B}} \circ {\phi}_{\text{fix}}, I_{\text{A}})+ \lambda\|\nabla({\phi}_{\text{mov}}\circ {\phi}_{\text{fix}})-I\|_F^2,
\end{equation}
where $\phi_{\text{A}}$ denotes the deformation from $I_{\text{A}}$ to $I_{\text{B}}$ and $\phi_{\text{A}}$ denotes the deformation from $I_{\text{B}}$ to $I_{\text{A}}$. The operator $\nabla$ denotes the Jacobian of a deformation field, and $I$ is the identity matrix. $\|\cdot\|_F^2$ denotes the square of the Frobenius norm of a matrix.

\subsection{Gradient Projection}
\begin{figure}[htbp]
    \centering
    \includegraphics[scale=0.75]{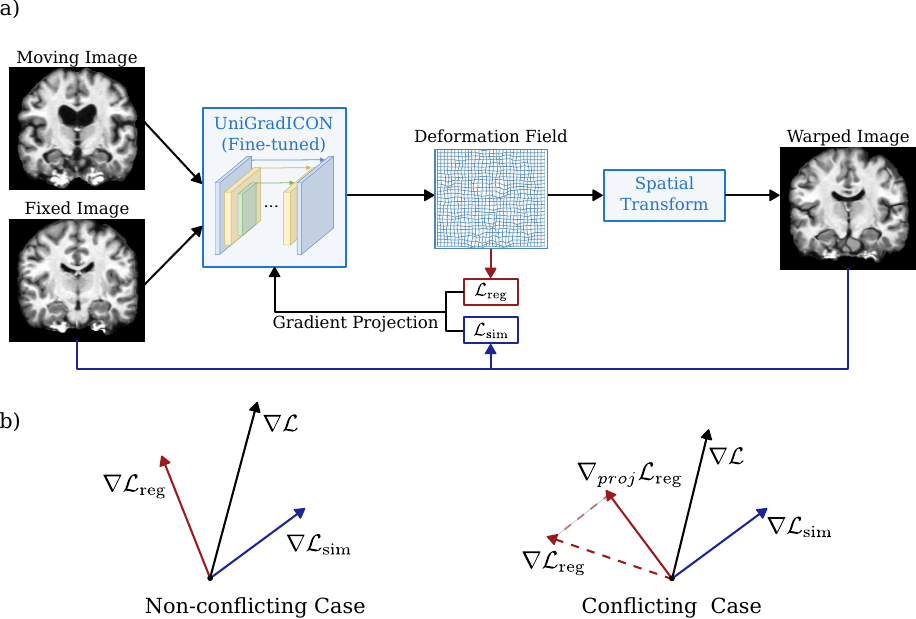}
    \caption{An illustrative pipeline of the proposed method. In (a), instead of taking the weighted sum of $\nabla\mathcal{L}_{\text{sim}}$ and $\nabla\mathcal{L}_{\text{reg}}$, we use the gradient projection as shown in (b). The gradients can be divided into two cases: non-conflicting and conflicting cases. In the non-conflicting case, the network update $\nabla\mathcal{L}$ is given by the vanilla gradient descent. In the conflicting case, $\nabla\mathcal{L}_{\text{reg}}$ is projected to the normal space of $\nabla\mathcal{L}_{\text{sim}}$, resulting in $\nabla_proj\mathcal{L}_{\text{reg}}$. Note that the choice of the gradient to be projected is \textit{random}. The loss hyperparameter $\lambda$ is omitted here without loss of generality.}
    \label{fig:enter-label}
\end{figure}

In MOO, multiple losses ($\mathcal{L}_{\text{sim}}$ and $\mathcal{L}_{\text{reg}}$) can sometimes conflict with each other, affecting the optimization. This can be reflected by the direction and magnitude of the loss gradients. To address this problem, we use the multitask learning gradient projection technique \cite{yu2020gradient}. The idea is to correct the conflicting direction of the loss gradient by \textit{randomly} projecting one gradient into the normal space of the other. We define the gradient of losses \textit{w.r.t.} model parameters $\theta$ as $\nabla\mathcal{L}_{\text{sim}}$ and $\nabla\mathcal{L}_{\text{reg}}$ for simplicity. We identify the gradient conflicting scenario by evaluating the cosine similarity between two gradients. Unlike mini-batch calculation \cite{yu2020gradient,dou2023gsmorph}, we focus on instance optimization in this work, thus reducing the scope from multi-task learning to multi-objective optimization. If the gradients are not conflicting, the update will remain unchanged compared with gradient-based methods. On the other hand, if the gradients are conflicting, one of the gradients will be \textit{randomly} projected to the normal space of the other one. For example, if $\nabla\mathcal{L}_{\text{sim}}$ is chosen to be corrected, the resulting  corrected gradient is calculated as
\begin{equation}
    \nabla\mathcal{L}_{\text{sim}} = \nabla\mathcal{L}_{\text{sim}} - \frac{\langle\nabla\mathcal{L}_{\text{sim}}, \nabla\mathcal{L}_{\text{reg}}\rangle}{\|\nabla\mathcal{L}_{\text{sim}}\|^2} \nabla\mathcal{L}_{\text{reg}},
\end{equation}
where $\langle\cdot,\cdot\rangle$ denotes the inner product of two vectors. An illustrative pipeline of our proposed method is shown in Fig. \ref{fig:enter-label}.

\section{Experiments}
\subsection{Data}
We conducted our experiments on the L2R Challenge LUMIR dataset \cite{marcus2007open,dufumier2022openbhb,taha2023magnetic}, which consists of T1-weighted brain MRI scans from 10 public datasets. The dataset includes 3384 subjects for training, 40 subjects for validation, and 590 subjects for testing. All images were resampled and cropped to focus on the region of interest, resulting in an image size of
$[160,224,192]$ with isotropic voxel spacing of $1\si{mm}^3$. To evaluate the performance of our method, we tested it in a local environment using 10 segmentation maps provided by the organizers, derived from the training subjects. These subjects were excluded from the training set to ensure unbiased evaluation. Additionally, we assessed the method on the validation set using the official platform. However, due to limitations on the number of uploads allowed, we report only a subset of the results from the validation split.

\subsection{Training Configurations}
We first further train the uniGradICON model with the training subjects. To fit the input requirement of the model, we rescale the images to a size of $[175,175,175]$. We use a batch size of 4 for training. We use Adam optimizer with an initial learning rate of $\eta = 5\times10^{-5}$ for 80 epochs. During instance optimization, we use the AMSGrad optimizer, initialized $\eta = 2\times10^{-5}$ with a weight decay of $10^{-3}$ for both vanilla IO and gradient projection. For all experiments, we set the balancing parameter $\lambda$ to be 1.5 as recommended by GradICON. We test the IO steps with two choices: 50 steps and 100 steps. 

Data augmentation is considered after 50 epochs of training, independently for the images in a single pair. The augmentation includes random rotation $[-10^{\circ}, +10^{\circ}]$, translation $[-5\si{mm},+5\si{mm}]$, scaling $[-3\%,+3\%]$, vertical flip, and $\gamma$ change $[-0.2,+0.2]$. The augmentation is performed with a probability of $50\%$, except the $\gamma$ change which is $10\%$.
\section{Results and Discussion}
We evaluated three approaches: direct inference (without IO), vanilla IO, and gradient projection IO with steps $n = 50, 100$. in the leave-out training set with segmentation labels. These evaluations were conducted on the leave-out training set, which included segmentation labels, and the results are summarized in Table \ref{tab:model ablation}. The metrics used for assessment included the Dice similarity coefficient, Hausdorff Distance (HD95), and Non-diffeomorphic volumes (NDV) \cite{liu2024finite}.

Due to upload limitations, we only report validation results for vanilla IO and gradient projection IO (steps $n=100$) in Table \ref{tab:comparison}. For the validation data, we also included the target registration error (TRE) in the annotated landmarks as an additional evaluation metric. We compared our methods with a competitive baseline model, TransMorph \cite{chen2022transmorph}, which does not use IO.

\begin{table}
    \caption{Comparison of IO settings on the leave-out training set with segmentation labels. IO-GP denotes the IO with the gradient projection.}
    \centering
    \begin{tabular}{p{3cm} m{2.55cm}<{\centering} m{2.55cm}<{\centering} m{2.55cm}<{\centering} m{2.55cm}<{\centering}}
        \toprule
        \multicolumn{1}{l}{Model Settings}  &Dice $\uparrow$ & {HD95 $\downarrow$}& {NDV} $(\si{10^{-4}})$ $\downarrow $\\ 
        \midrule
        Direct Inference & $74.57\pm2.65$  & $3.58\pm0.30$  & $0.106\pm0.136$ \\
        \midrule
        IO ($n=50$) & $75.23 \pm 2.55$ & $3.51 \pm 0.30$ & $0.405 \pm 0.343$ \\
        IO-GP ($n=50$) & \textbf{$75.26 \pm 2.56$} & \textbf{$3.50 \pm 0.29$} & $0.400 \pm 0.351$ \\
        \midrule
        IO ($n=100$) & $75.39 \pm 2.55$ & $3.51 \pm 0.29$ & $0.678 \pm 0.506$ \\
    IO-GP ($n=100$) & $75.40 \pm 2.54$ & $3.51 \pm 0.30$ & $0.625 \pm 0.536$ \\
        \bottomrule
    \end{tabular}

    \label{tab:model ablation}
\end{table}

\begin{table}
    \caption{Comparison of IO settings on the validation set with segmentation labels. IO-GP denotes the IO with the gradient projection. For simplicity, we only report the standard error of Dice.}
    \centering
    \begin{tabular}{p{3cm} m{2.2cm}<{\centering} m{2.2cm}<{\centering} m{1.8cm}<{\centering} m{1.8cm}<{\centering}<{\centering} m{1.8cm}<{\centering}}
        \toprule
        \multicolumn{1}{l}{Model Settings}  &Dice $\uparrow$ & {HD95 $\downarrow$}& {NDV} $(\%)$ $\downarrow $ & {TRE} (\si{mm}) \\ 
        \midrule
        TransMorph & $75.94\pm3.08$  & $3.51$  & $0.351$ &2.4225\\
        \midrule
        IO ($n=100$) & $76.54 \pm 3.43$ & $3.34$ & $0.001 $ &2.3375  \\
    IO-GP ($n=100$) & $75.40 \pm 2.54$ & $3.33$ & $0.001 $ &2.3143\\
        \bottomrule
    \end{tabular}

    \label{tab:comparison}
\end{table}

\section{Discussion and Conclusion}
Our findings indicate that gradient projection IO achieved the best results across all evaluated criteria on the validation set, demonstrating superior performance compared to both vanilla IO and the TransMorph baseline. However, we notice that the performance gain is relatively marginal compared with vanilla IO when steps grow. Future studies could include a more comprehensive study on the impact of different optimizers and the criteria for determining stopping points for IO.

In conclusion, inspired by advances in multi-objective optimization, we have proposed a gradient projection method to refine the gradient-based updates of a pre-trained registration network. This method involves projecting gradients via a simple inner product operation, offering a lightweight solution that significantly enhances registration performance. The simplicity of this approach makes it efficient while effectively addressing conflicts between multiple optimization objectives even more than two as shown in the paper. Overall, our approach provides a practical enhancement to image registration tasks, demonstrating the potential for improved outcomes in complex medical imaging scenarios.
\bibliographystyle{splncs04}
\bibliography{bib}
\end{document}